\pdfoutput=1

\RequirePackage{snapshot}
\documentclass[11pt]{article}
\usepackage{booktabs}
\usepackage{graphicx}
\usepackage{caption}
\usepackage{subcaption}
\usepackage{amssymb}
\usepackage{multirow}
\usepackage{listings}
\usepackage{amsmath}
\usepackage[hidelinks]{hyperref}
\usepackage[noabbrev,capitalize]{cleveref}
\usepackage[normalem]{ulem}
\usepackage{enumitem}
\usepackage{acl}
\usepackage{times}
\usepackage{latexsym}
\usepackage{fancyvrb}
\usepackage[T1]{fontenc}
\usepackage{listings}

\usepackage[utf8]{inputenc}

\usepackage{microtype}

\usepackage{bbm}
\newcommand{\RN}[1]{%
  \textup{\uppercase\expandafter{\romannumeral#1}}%
}

\newcommand{\data}[1]{X_{#1}}
\newcommand{\model}{M}
\newcommand{\modelresponse}[1]{r_{#1}}
\newcommand{\critique}[1]{c_{#1}}
\newcommand{\instruction}[1]{I_{#1}}

\newcommand{\protocol}{SR$^2$V}

\usepackage{array}
\newcolumntype{C}[1]{>{\centering\arraybackslash}m{#1}}
\newcolumntype{H}{>{\setbox0=\hbox\bgroup}c<{\egroup}@{}}

\definecolor{applegreen}{rgb}{0.55, 0.71, 0.0}
\usepackage{contour}

\usepackage{clipboard}
\usepackage{stmaryrd}

\usepackage{csquotes}

\usepackage{array}
\newlength{\extramargin}
\usepackage{tabularx}

\title{When Hindsight is Not 20/20: \\Testing Limits on Reflective Thinking in Large Language Models}
\author{Yanhong Li$^{1,*}$, Chenghao Yang$^{1,*}$, Allyson Ettinger$^2$  \\
  $^1$University of Chicago 
  $^2$Allen Institute for AI \\
  \texttt{\{yanhongli, chenghao\}@uchicago.edu} \\
  \texttt{allysone@allenai.org} \\
  }

\begin{document}
\maketitle
\def\thefootnote{$*$}\footnotetext{Equal Contribution.}\def\thefootnote{\arabic{footnote}}
\def\thefootnote{\arabic{footnote}}
\begin{abstract}
Recent studies suggest that self-reflective prompting can significantly enhance the reasoning capabilities of Large Language Models (LLMs). However, the use of external feedback as a stop criterion raises doubts about the true extent of LLMs' ability to emulate human-like self-reflection. In this paper, we set out to clarify these capabilities under a more stringent evaluation setting in which we disallow any kind of external feedback. Our findings under this setting show a split: while self-reflection enhances performance in TruthfulQA, it adversely affects results in HotpotQA.
We conduct follow-up analyses to clarify the contributing factors in these patterns, and find that the influence of self-reflection is impacted both by reliability of accuracy in models' initial responses, and by overall question difficulty: specifically, self-reflection shows the most benefit when models are less likely to be correct initially, and when overall question difficulty is higher. We also find that self-reflection reduces tendency toward majority voting. Based on our findings, we propose guidelines for decisions on when to implement self-reflection. We release the codebase for reproducing our experiments at \url{https://github.com/yanhong-lbh/LLM-SelfReflection-Eval}.
\end{abstract}

\section{Introduction}

Large Language Models (LLMs) have shown impressive performance in generating human-like text (e.g., ChatGPT \cite{chatgpt}), and recent works demonstrate that we can further prompt LLMs to reflect on their own outputs to improve their capabilities on complicated reasoning, programming and planning tasks~\citep{huang2022large, kim2023language, madaan2023selfrefine, Shinn2023ReflexionLA, chen2023teaching, wang2023enable} and also improve their alignment with human values (e.g., less harmful and more helpful)~\citep{bai2022constitutional, ganguli2023capacity}.\footnote{Various terms like ``self-reflection'', ``self-refine'',  ``self-correction'', and ``self-improvement'' describe these introspective behaviors. For clarity and consistency, we will exclusively use ``self-reflection'' in this paper.} However, 
\citet{huang2023large} find that performance gains associated with self-reflection may be due to implicit usage of external feedback as a stop criterion, as well as overly-engineered prompts that bias the model outputs, casting doubt on the true effectiveness of self-reflection. 

To verify the extent to which LLMs can truly reflect on their outputs, we take a more stringent evaluation approach: in addition to excluding external feedback~\citep{huang2023large}, we also disallow multi-round iterative prompting, which can hint to the model that its prior response is incorrect. Instead, we sample multiple model responses given a prompt, and ask the model to self-reflect on these candidate outputs. With this \emph{single-round testing}, 
we can zero in on the model's ability to use self-reflection without implicit hints about whether a given response candidate is correct or incorrect.

\begin{figure}[tbp!]
\centering
\small
  \includegraphics[width=0.47\textwidth]{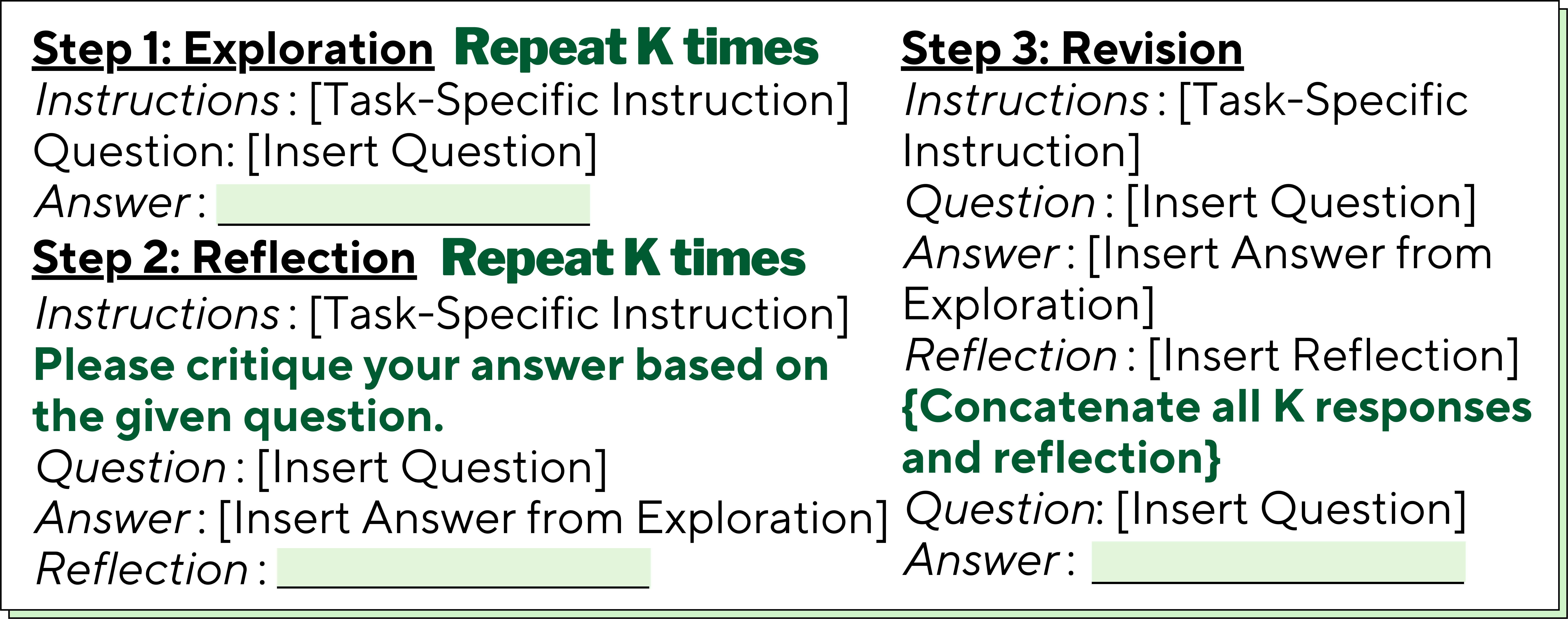}
      \vspace{-1mm}
  \caption{Example of Self-Reflection Prompting 
  }
    \label{fig:PromptingExample}
        \vspace{-4mm}
\end{figure}

Our experiments show that, in a case study with ChatGPT on different QA datasets, self-reflection in our setting yields mixed results. Specifically, self-reflection improves performance on TruthfulQA \citep{lin2022truthfulqa}, but decreases model performance in HotpotQA \citep{yang2018hotpotqa}.
Through follow-up analyses, we identify that the effectiveness of self-reflection strongly depends on the confidence in accuracy of the model's initial responses, as well as overall question difficulty as judged by humans: when the model is reliably giving correct answers from the start, self-reflection is more often harmful---however, on questions of greater difficulty, self-reflection is beneficial even when a decent percent of initial model responses are correct. We also find that self-reflection reduces model tendency toward majority voting, suggesting more sophisticated decision-making (albeit sometimes resulting in lower accuracy). 
Based on our findings, we propose a practical guideline for users to decide when to use self-reflection.

\section{Self-Reflection Prompting}

To focus on evaluating intrinsic reflective thinking capability, we adopt the following 
evaluation setting: 
in addition to the \citet{huang2023large} protocol of excluding external feedback and prompt optimization, we additionally disallow \emph{iterative prompting}, which samples new responses based on previous responses, creating an implicit hint to bias the model behavior \citep{huang2023large}.\footnote{We present a performance comparison between iterative prompting and non-iterative prompting in \cref{app:conditional_prompting}.} We call our approach \emph{Single-Round Self-Reflection Verification (\protocol)}. We evaluate LLMs' reflective thinking capability using the following simple three-stage format: 
1) \emph{Exploration:} Given an input $\data{}$, we prompt LLM $\model$ to generate $K$ candidate responses 
$\modelresponse{j} \sim P_\model(\modelresponse{j} | \data{}, \instruction{\text{Exploration}}), 1\leq j \leq K$ with instruction $\instruction{\text{Exploration}}$. Note that this generation of candidate responses differs from iterative prompting because each response is sampled without conditioning on any other candidate responses.
2) \emph{Reflection:} For each response $\modelresponse{j}$, we prompt $\model$ with the concatenated input $[\data{}; \modelresponse{j}]$ to generate a self-critique $\critique{j} \sim P_\model\left(\critique{j} | [\data{}; \modelresponse{j}], \instruction{\text{Reflection}}\right)$ with another instruction $\instruction{\text{Reflection}}$.
3) \emph{Revision:} 
We concatenate the $K$ response-reflection pairs into a new input and prompt $\model$ to generate an improved output. An illustration of this procedure is shown in \cref{fig:PromptingExample}.

\section{Preliminary Study: Does Self-Reflection Prompting Work Under \protocol?}

We follow previous works~\cite{bai2022constitutional, Shinn2023ReflexionLA, huang2023large} 
in using two representative datasets, TruthfulQA and HotpotQA, to verify the effectiveness of self-reflection under \protocol. TruthfulQA is designed to evaluate the truthfulness of LMs' responses, 
while HotpotQA focuses on multi-hop reasoning tasks, aimed at requiring complex reasoning capabilities. 

\paragraph{Experiment Setup} For these experiments we set $K=4$, 
and we prompt ChatGPT-3.5 (``gpt-3.5-turbo-16k-0613'')
with the questions from each dataset.\footnote{The 16k variant is chosen to accommodate responses and reflection pairs that exceed the standard 4096 token limit, particularly in detailed experiments of \cref{sec:artificial_response}.} Our full process for making these API calls is presented in \cref{app:api_call_example}, and all prompt templates used can be found in \cref{app: prelim-prompt-templates}. We also extend our experiments to LLaMA-2~\cite{touvron2023llama} and Mixtral~\cite{jiang2024mixtral}, finding similar results to ChatGPT-3.5---we present results and discussion for LLaMA-2 and Mixtral in \cref{app:llama2_results} and \cref{app:mixtral_results}. 

For TruthfulQA we evaluate automatically (see details in \cref{app:truthfulqa_evaluation}).  For HotpotQA, we find that 
traditional exact match often unfairly assigns a score of $0$ for semantically correct model responses; 
therefore, we manually assess $1,000$ randomly chosen HotpotQA instances to check the model’s answers against references.  

To isolate the specific effect of the generated reflections, we also include an \textbf{exploration-only} baseline, in which we retain the Exploration stage but remove the Reflection component, and only concatenate the candidate model responses in the Revision prompt.\footnote{
The exploration-only baseline can be viewed as one implementation of (universal)  self-consistency prompting~\citep{wang2023selfconsistency,chen2023universal}. 
Rather than applying majority voting directly to the outputs, this method involves inputting these outputs back into the model for aggregation. As we'll explore in \cref{sec:mv}, we also find the model predominantly engages in a form of majority voting in this process.
} 

\begin{table}[t!]
\centering
\resizebox{0.49\textwidth}{!}{%
\begin{tabular}{@{}p{2cm}p{2cm}p{2cm}p{2cm}@{}}
\toprule
Metric & {Standard Prompting} & Exploration-Only & Self-Reflection \\
\midrule
\multicolumn{4}{c}{TruthfulQA} \\
Rouge-1 & $57.5 \pm 1.1$ & $57.2$ & $\textbf{60.8}$ \\
BLEURT & $66.8 \pm 1.9$ & $60.7$ & $\textbf{72.8}$ \\
\midrule
\multicolumn{4}{c}{HotpotQA} \\
Accuracy* & $80.3 \pm 0.5$ & $\textbf{80.8}$ & $76.2$ \\
EM & $\mathbf{50.5 \pm 0.4}$ & $47.3$ & $37.0$ \\
\bottomrule
\end{tabular}
}
\caption{Self-reflection \protocol{} experiment results on QA datasets. Bold-facing indicates the best-performing method under each metric. *Evaluated manually. 
}
\label{tab: merged_res}
\vspace{-5pt}
\end{table}

\paragraph{Observations} Results are shown in~\cref{tab: merged_res}.
In TruthfulQA, 
we see that using self-reflection achieves significantly better performance than either the exploration-only baseline or standard prompting. 
This finding is consistent with the observation of~\citet{bai2022constitutional} that LLMs' self-evaluation (in the form of reflection) can help to produce more factual outputs. 
However, we see that on HotpotQA, accuracy when using self-reflection is about $4\%$ worse compared to both the exploration-only baseline and standard prompting. These results suggest that self-reflection may in fact harm performance in multi-hop reasoning tasks. This aligns with the self-reflection limitations found in~\citet{huang2023large}, and verifies that these limitations also extend to our more stringent evaluation setting, but presents a more complicated picture with the continued effectiveness of self-reflection on TruthfulQA under this setting.

\begin{figure}{}
\centering
\small
\includegraphics[width=0.4\textwidth]{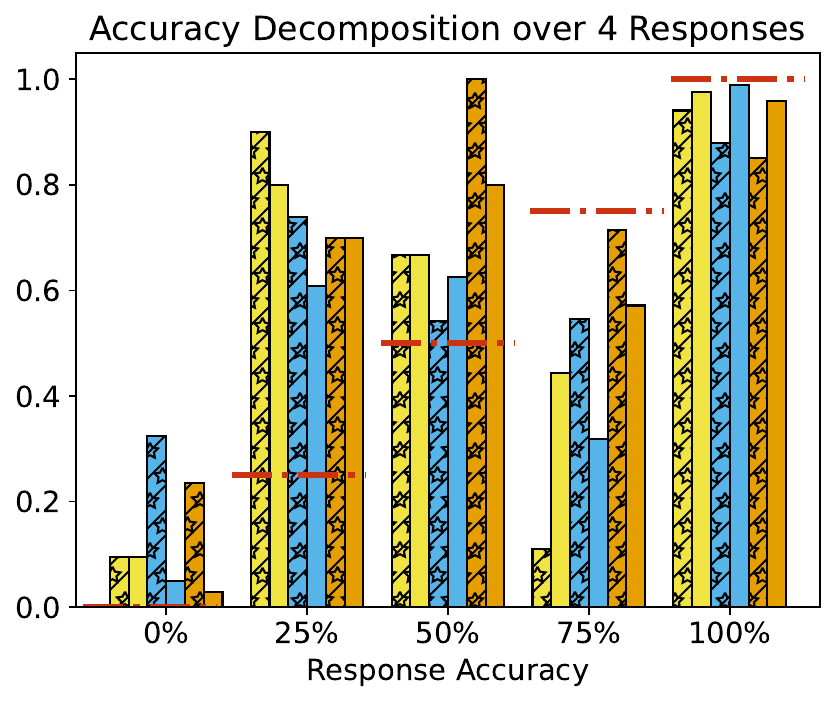}
  % \vspace{-2mm}

\includegraphics[width=0.4\textwidth]{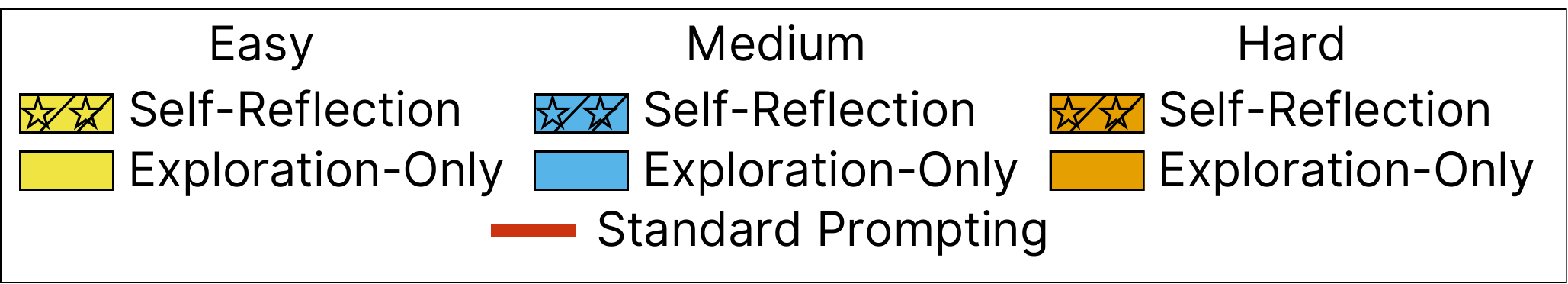}
  % \vspace{-2mm}

  \caption{Performance Decomposition on Question Difficulty and Response Accuracy. 
  }

  \label{fig:perf_decomp_4response_natural}
  \vspace{-5mm}
\end{figure}

\section{Why Self-Reflection May Not Work?}
\label{sec:init_error_analysis}
To better understand these patterns, we conduct an error analysis drawing inspiration from the reflection conceptual model in psychology~\cite{hommel2023reflection}. We hypothesize that two key factors influence self-reflection's efficacy: 1) the objective \textbf{question difficulty} (quantifiable based on human annotations), 
and 2) the \textbf{model's comprehension quality} (quantifiable based on the proportion of correct responses). 
Following this framework, we can predict that if a question is above average in human-annotated difficulty, self-reflection may be of greater benefit. Similarly, if the model already has a strong grasp of the question, it may not benefit as much from self-reflection. 

To test these hypotheses, we break down model performance based on levels of question difficulty and model comprehension. We focus our analysis on HotpotQA, as this is the dataset on which we observe significant detrimental effects of applying self-reflection prompting. Additionally, this dataset contains annotated human judgments of question difficulty, and enables a clearly-defined notion of accuracy. We use these human difficulty annotations for our measure of question difficulty, and for model comprehension we use Response Accuracy (RA): the proportion of correct answers among the K candidate model responses sampled during Exploration. 

 The broken-down results are shown in \cref{fig:perf_decomp_4response_natural}. The results show an interaction between our two variables. For questions judged by humans as Easy, self-reflection shows a benefit only when the model's candidate responses are mostly---but not all---incorrect, with self-reflection otherwise having negligible or negative effects on performance. For questions judged as Medium, there is a more even split: when most or all of the model's candidate responses are wrong, self-reflection is beneficial, but when half or more of the responses are correct, self-reflection is often harmful---with the notable exception of the 75\% RA bin. A similar pattern is seen for questions judged as Hard, though for this category self-reflection is more consistently beneficial through the 75\% RA bin, showing harm to performance only when all candidate model responses are already correct.

\begin{figure}[t!]
\centering
\small
  \includegraphics[width=0.482\textwidth]{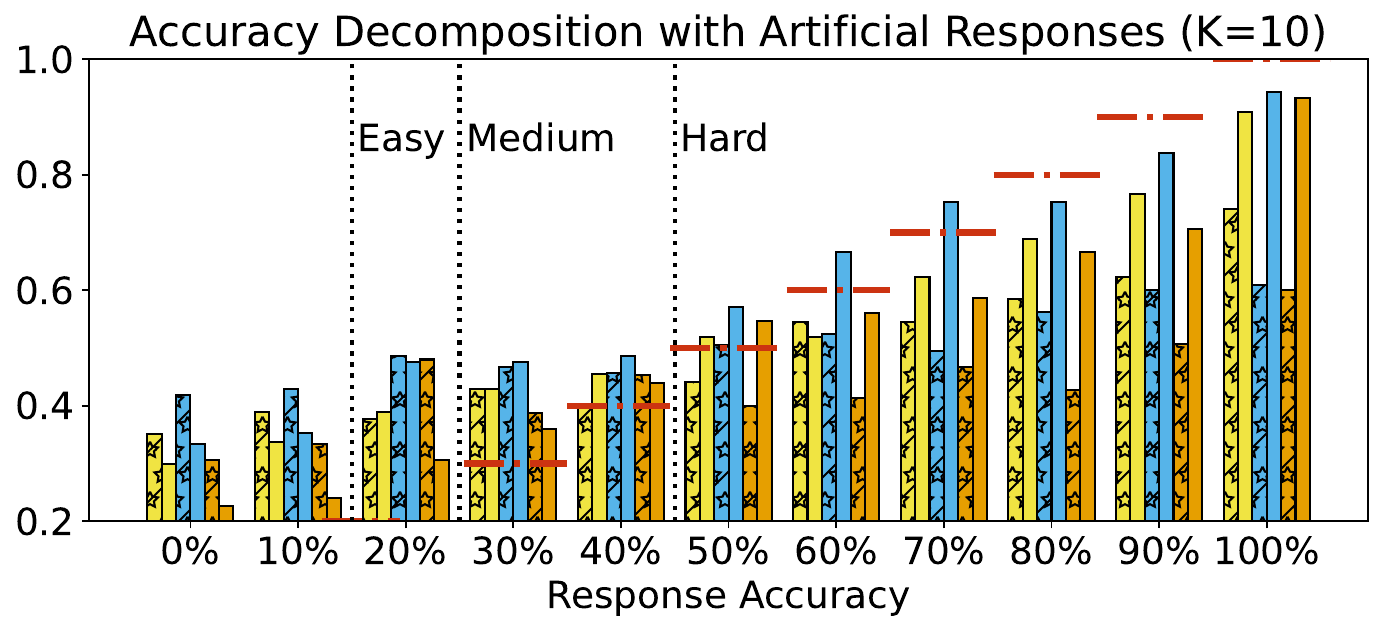}
    \includegraphics[width=0.4\textwidth]{figures/legend.pdf}
  \vspace{-2mm}
  \caption{Performance Decomposition on Question Difficulty and Response Accuracy (Artificial Responses). Dotted lines show ``turning points'' at which reflection loses effectiveness, for Easy/Medium/Hard questions. }
  \label{fig:10level_fake}
  \vspace{-5mm}
\end{figure}

\section{Error Analysis via Artificial Response}
\label{sec:artificial_response}
The above analysis suggests an interaction between difficulty and comprehension variables in effectiveness of self-reflection---however, our ability to disentangle these effects is limited by imbalanced distribution of model comprehension relative to question difficulty. To assess the interaction more thoroughly, we simulate model ``mis-comprehension'' across a wider range of question difficulties, by sampling model responses to minimally edited versions of the prompts, and then pairing these responses with the original prompts when eliciting self-reflection. This allows us to increase the number of incorrect candidate responses, and thus to more evenly distribute RA levels across human difficulty levels. More details on this simulation process can be found in \cref{app:artificial_response_generation}.

For this experiment, we generate K $=10$ candidate responses per question, with a mix of synthetic pairings and real pairings.\footnote{We also plot the performance decomposition over K=4 artificial responses in \cref{app:fig_perf_decomp_art_resp}.} 
Results are shown in
\cref{fig:10level_fake}. We see that the benefits of self-reflection are now limited to the lowest RA levels, and there is also now a clearer shift from beneficial to harmful effects of self-reflection as RA increases. We also see that the interaction with question difficulty remains: the turning point from beneficial to harmful falls around 50\% RA for Hard questions, 30\% for Medium questions, and 20\% for Easy questions. Overall, this indicates that a major contributor to the effectiveness of self-reflection is the confidence of model accuracy on the question---if the model is reliably correct on initial responses, self-reflection tends to be harmful. However, this effect is further modulated by overall question difficulty: the benefits of self-reflection persist to higher levels of response accuracy if the questions are more difficult based on human judgment.

Though TruthfulQA is not as conducive to exact quantification of our variables, based on these results we can now speculate that the effectiveness of self-reflection on that dataset may be attributable to lower rate of good initial model responses, and potentially also higher overall question difficulty.

\begin{figure}[htbp!]
\centering
  \includegraphics[width=0.4\textwidth]{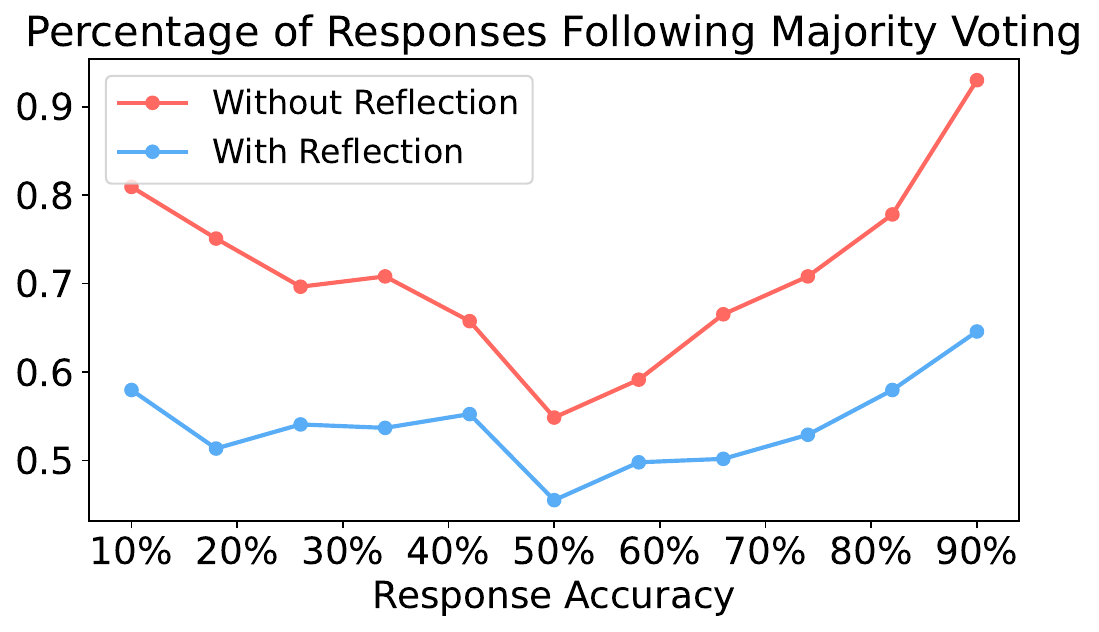}
  \caption{Majority Voting Analysis}
    \label{majority_voting}
    \vspace{-4mm}
\end{figure}

\section{Effects on majority voting}\label{sec:mv}
A natural question to ask at this point is to what extent the effect of RA is due to the model employing majority voting on the candidate responses.
In \cref{majority_voting} we plot the percentage of items in which the model's output is consistent with majority voting, at different RA levels (computed at $K=10$ including artificially generated responses), both with and without self-reflection. The plot shows that without self-reflection, the tendency to give answers consistent with majority voting is strong and closely correlated with the strength of the accuracy trend (i.e., more majority voting when most candidate responses are either correct or incorrect, and less majority voting when candidates are more mixed). However, \emph{with} self-reflection the tendency to align with majority voting is significantly reduced across RA levels, suggesting that self-reflection does encourage more sophisticated decision strategies (even if in the case of higher RA levels, this in fact has a harmful effect on accuracy).

\begin{figure}[tbp!]
\centering
\small
  \includegraphics[width=0.42\textwidth]{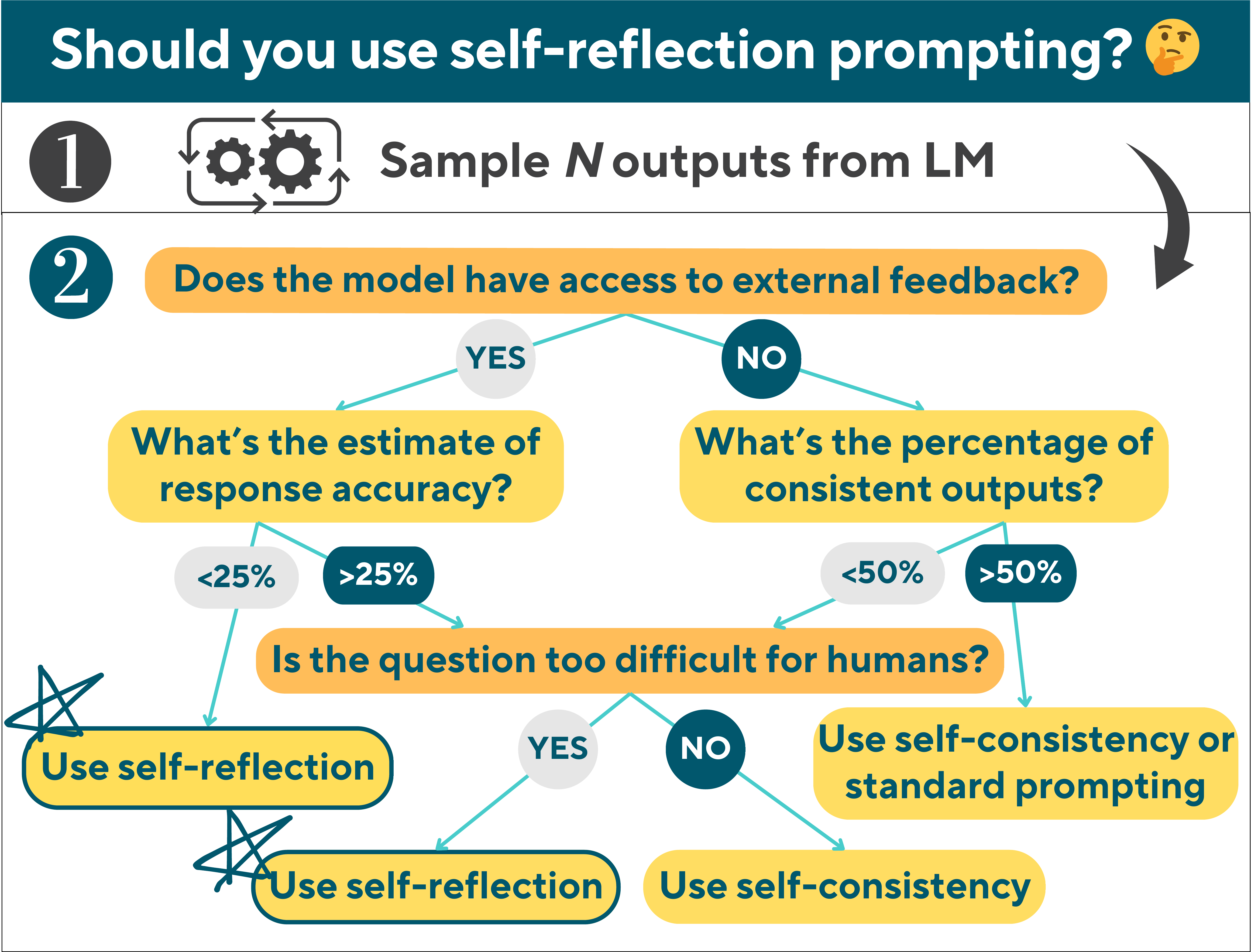}
  \caption{Proposed guide for using Self-Reflection. 
  } 
    \label{fig:guideline_self_reflection}
\end{figure}

\section{Discussion} 
Our analyses above have found that self-reflection benefits are limited to cases in which model accuracy is unreliable on initial responses, though benefits are more persistent for harder questions.
Based on these findings, we propose a set of guidelines for determining when to implement self-reflection in practical applications, for a given request or prompt. The core principle involves basing decisions on estimated RA and question difficulty, and these guidelines can be applied by simply sampling responses for the target question or prompt. First, if external tools or certain access to ground truth answers are available such that RA can be reliably estimated, then self-reflection should be used when RA levels are low. Next, if difficulty annotations/subjective difficulty judgements are available, self-reflection can also be promising when RA levels are intermediate and question difficulty is high. If RA cannot be estimated, response consistency can be used as a proxy: if responses are highly consistent, self-reflection may be unlikely to provide benefit. If consistency is low, then self-reflection may be beneficial, especially for questions of higher difficulty.
An illustration of these guidelines is in \cref{fig:guideline_self_reflection}. 

\section{Conclusion}
In this paper, we evaluate ChatGPT's self-reflective capabilities under a stringent single-round multi-response evaluation setting. We find mixed results, and further analysis shows that the effectiveness of self-reflection is impacted both by question difficulty and by model response accuracy level: benefits of self-reflection are mostly limited to cases in which the model's initial responses are unreliable in accuracy, but with more persistent benefits for harder questions. Additionally, we find that self-reflection reduces the model's tendency for majority voting. We propose guidelines for when to use self-reflection, and we look forward to work further exploring impacts on self-reflection, and further refining these guidelines. 

\section*{Acknowledgements}
We are grateful for the insightful discussion with Xinyun Chen (Google) and Jie Huang (UIUC) at the early stage of this work (names are not listed in particular order). We also thank the anonymous NAACL reviewers and chairs for providing insightful and constructive feedback to make this work more solid. 

\section*{Limitations}

In this work, we adopt a stringent evaluation strategy to test the effectiveness of self-reflective abilities of LLMs. One limitation is that our experiments reported in the main text are based on a single snapshot of the ChatGPT model (gpt-3.5-turbo-16k-0613). We focus on ChatGPT because it is a state-of-the-art chat model, allowing us to make our results directly comparable with previous work---and we limit to this particular version of ChatGPT to ensure that results will not be affected by model updates. However, the assessment of self-reflection may vary between different versions of ChatGPT, as well as between ChatGPT and other LLMs. We do verify our experimental results on other open language models including LLaMA-2~\cite{touvron2023llama} and Mixtral~\cite{jiang2024mixtral} in \cref{app:llama2_results} and \cref{app:mixtral_results}, respectively. While we find that our conclusions can be extended to these models, due to budget limitations we leave more extensive evaluation over other popular proprietary and open models for future works.  

Our experiments also use only two datasets for evaluating reflective ability. We chose these two datasets for a focused study covering two very different QA domains, but we look forward to future work further extending these types of analyses to a broader collection of datasets.

We conducted an artificial response experiment in \cref{sec:artificial_response} to simulate the real output distribution of the language model. This is a rough estimate of ChatGPT's actual output distribution. As we sampled ten fake responses from the language model, it is impossible to cover all possible cases of outputs, and there might be bias in the sample distribution. Future work could try generating a higher number of fake responses to obtain a more accurate distribution of the model.

Finally, although RA proves a valuable metric for determining the utility of self-reflection, its reliance on access to ground truth undermines its practical use. An initial attempt to use GPT-4 to produce an estimate of RA yielded unsatisfactory results (detailed in \cref{app:challenge_predict_correctness_margin}). Further examination of this topic is reserved for future research. 

\bibliography{main}
\bibliographystyle{acl_natbib}
\appendix
\section{Accuracy Decomposition over 4 responses}
\label{app:fig_perf_decomp_art_resp}
\begin{figure}[h]
\centering
  \includegraphics[width=0.4\textwidth]{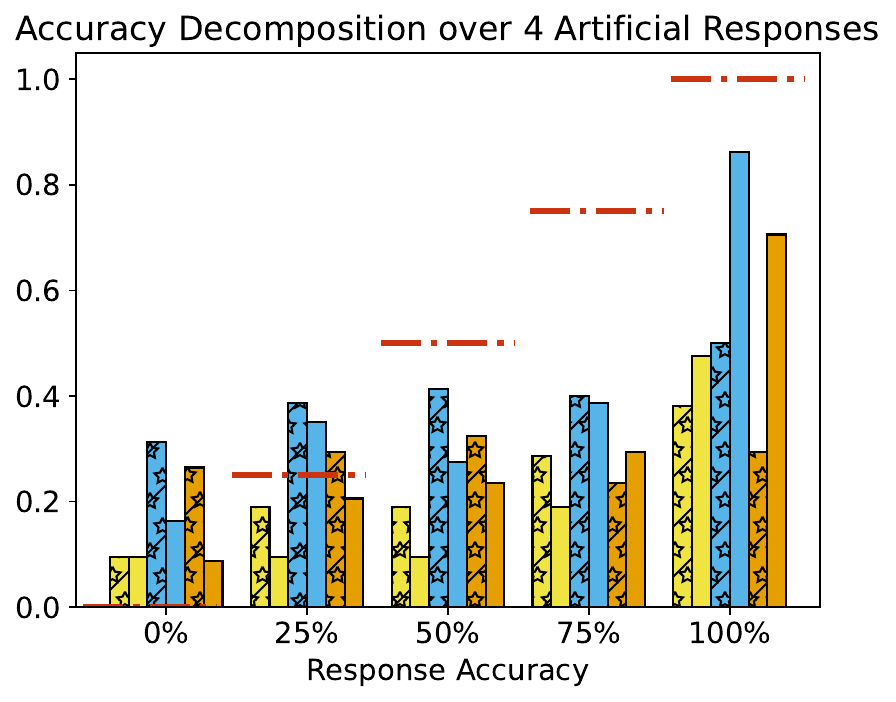}
  \includegraphics[width=0.4\textwidth]{figures/legend.pdf}
  \caption{Accuracy vs. Correctness Margin for each artificial response 
  }
  \label{fig:4level_fake}
\end{figure}

See \cref{fig:4level_fake} for accuracy decomposition over 4 responses using ChatGPT.

\section{Artificial Response Generation}
\label{app:artificial_response_generation}
\begin{figure}[htbp!]
\centering
  \includegraphics[width=0.42\textwidth]{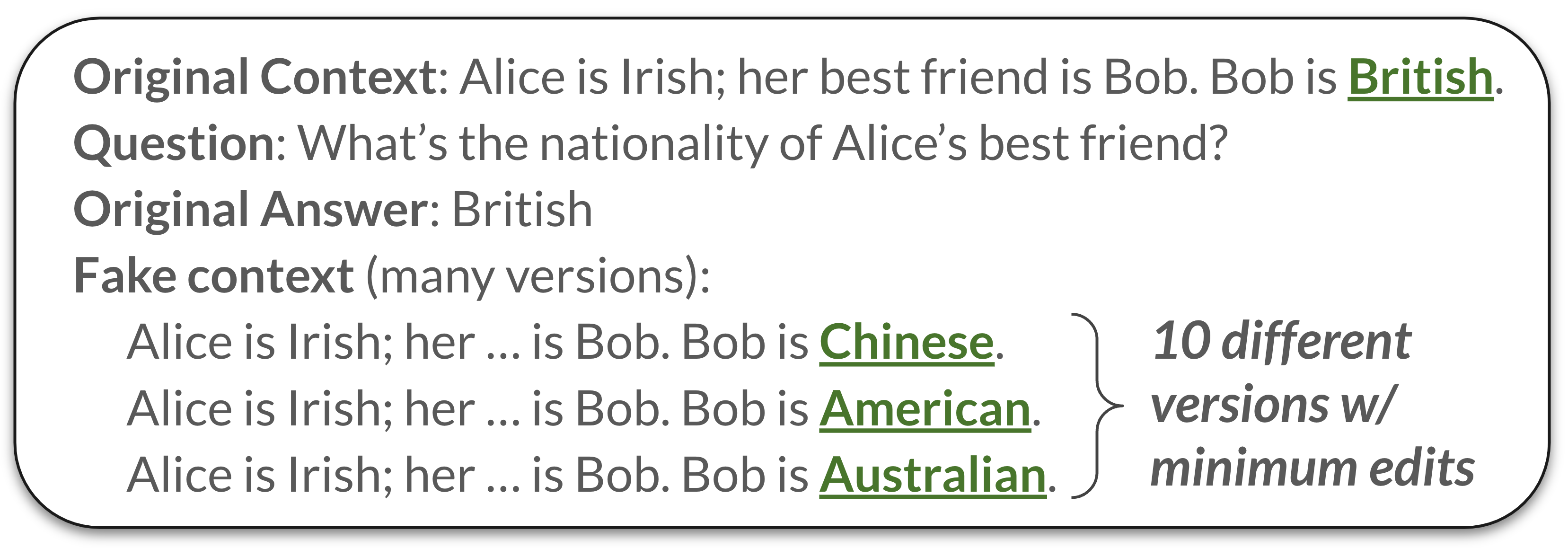}
  \vspace{-3mm}
  \caption{Synthesized Artificial Contexts Example 
  }
    \label{fig:FakeContext}
    \vspace{-5mm}
\end{figure}

We do artificial response generation by prompting ChatGPT to edit the context used in HotpotQA. 
Specifically, the following steps were adopted:
    1) For chosen questions, perform a simple perturbation on the context (e.g., entity replacement). An example is shown in \cref{fig:FakeContext}.
    2) Manually inspect some samples to ensure minimal edits and answerability.
    3) Prompt the model to regenerate responses and reflections based on the altered context. 
In this way, we are simulating scenarios where the model doesn't comprehend the context perfectly. \footnote{While directly editing outputs to create correct or incorrect answers is an option, we avoid this to ensure the results reflect the model's natural response distribution.}

Here is an example for how we modify the context:

\textbf{Original question}: What nationality was James Henry Miller's wife?

\textbf{Original context}: ... Ewan MacColl: James Henry Miller (25 January 1915 – 22 October 1989), better known by his stage name Ewan MacColl, was an \text{\color{blue} English} folk singer, songwriter, 
\text{\color{blue} communist}, labour activist, actor, poet, playwright and record producer. Peggy Seeger: Margaret "Peggy" Seeger (born June 17, 1935) is an American \text{\color{blue} folksinger}. She is also well known in \text{\color{blue} Britain}, where she has lived for more than 30 years, and was married to \text{\color{blue} the singer and songwriter}  Ewan MacColl until his death in 1989. ...

\textbf{Fake context 1}: ... Ewan MacColl: James Henry Miller (25 January 1915 – 22 October 1989), better known by his stage name Ewan MacColl, was a \text{\color{blue} Scottish} folk singer, songwriter, \text{\color{blue} capitalist}, labour activist, actor, poet, playwright and record producer.. Peggy Seeger: Margaret "Peggy" Seeger (born June 17, 1935) is an American \text{\color{blue} country} singer. She is also well known in \text{\color{blue} France}, where she has lived for more than 30 years, and was married to \text{\color{blue} the actor and playwright} Ewan MacColl until his death in 1989. ...

\textbf{Fake context 2}: ... Ewan MacColl: James Henry Miller (25 January 1915 – 22 October 1989), better known by his stage name Ewan MacColl, was an \text{\color{blue} Australian} folk singer, songwriter, \text{\color{blue} conservative}, labour activist, actor, poet, playwright and record producer. Peggy Seeger: Margaret "Peggy" Seeger (born June 17, 1935) is a \text{\color{blue}British pop} singer. She is also well known in \text{\color{blue}Germany}, where she has lived for more than 30 years, and was married to \text{\color{blue}the musician and producer} Ewan MacColl until his death in 1989. ...

\textbf{Fake context 3}: ... Ewan MacColl: James Henry Miller (25 January 1915 – 22 October 1989), better known by his stage name Ewan MacColl, was a \text{\color{blue} Canadian} folk singer, songwriter, \text{\color{blue} anarchist}, labour activist, actor, poet, playwright and record producer. Peggy Seeger: Margaret "Peggy" Seeger (born June 17, 1935) is an \text{\color{blue} American rapper}. She is also well known in \text{\color{blue}Spain}, where she has lived for more than 30 years, and was married to \text{\color{blue}the actor and politician} Ewan MacColl until his death in 1989. ...

\textbf{Fake context 4}: ... Ewan MacColl: James Henry Miller (25 January 1915 – 22 October 1989), better known by his stage name Ewan MacColl, was an \text{\color{blue}Irish} folk singer, songwriter, \text{\color{blue}monarchist}, labour activist, actor, poet, playwright and record producer. Peggy Seeger: Margaret "Peggy" Seeger (born June 17, 1935) is a \text{\color{blue}French jazz singer}. She is also well known in \text{\color{blue}Italy}, where she has lived for more than 30 years, and was married to \text{\color{blue}the artist and filmmaker} Ewan MacColl until his death in 1989. ...

\begin{table}[ht!]
\centering
\small
\resizebox{0.48\textwidth}{!}{%
\begin{tabular}{@{}lccc@{}}
\toprule
Metric & {Standard Prompting} & Exploration-Only & Self-Reflection \\
\midrule
\multicolumn{4}{c}{TruthfulQA} \\
Rouge-1 & $57.5 \pm 1.1$ & $55.1$ & $\textbf{59.0}$ \\
BLEURT & $66.8 \pm 1.9$ & $70.1$ & $\textbf{72.9}$ \\
\midrule
\multicolumn{4}{c}{HotpotQA} \\
Accuracy & $80.2 \pm 0.4$ & $69.7$ & $71.9$ \\
\bottomrule
\end{tabular}
}
\caption{Self-Reflection experiment results using iterative prompting. Bold-faced numbers at each row indicate the best-performing method under each metric. }
\label{tab: conditional_prompting}
\vspace{-10pt}
\end{table}

\section{Conditional Prompting Results}
\label{app:conditional_prompting}
We demonstrate the conditional prompting results in \cref{tab: conditional_prompting}. Comparing the results in \cref{tab: merged_res} and \cref{tab: conditional_prompting}, we can see that there is no significant difference between these parallel prompting and conditional prompting. To avoid the implicit bias introduced by conditional prompting, as \citet{huang2023large} point out, we stick to parallel prompting to conduct our evaluation on self-reflective thinking capability. 

\section{Evaluation details for TruthfulQA}
\label{app:truthfulqa_evaluation}
We use the generation setting of TruthfulQA, which evaluates by comparing how closely the model's responses match a preferred reference versus an undesired one
We follow~\cite{lin2022truthfulqa} to use Rouge-1~\cite{lin-2004-rouge} and BLEURT~\cite{sellam-etal-2020-bleurt} for similarity computation.

\section{Prompts used in Experiment}
\label{app: prelim-prompt-templates}

\subsection{TruthfulQA: Standard Prompt}
\begin{small}
\begin{lstlisting}
messages=[
    {"role": "user", 
    "content": question}
]
\end{lstlisting}
\end{small}

\subsection{TruthfulQA: Response Critique Prompt}
\begin{small}
\begin{lstlisting}
messages=[
    {"role": "system", 
    "content": "You are a helpful 
    assistant."},
    {"role": "user", 
    "content": question},
    {"role": "assistant", 
    "content": response},
    {"role": "user", 
    "content": "Could you critique 
    your last response?"}
]
\end{lstlisting}
\end{small}
\subsection{TruthfulQA: Response Without Reflection}
\begin{small}
\begin{lstlisting}
messages=[
    {"role": "system", 
    "content": "You are a helpful 
    assistant."},
    {"role": "user", 
    "content": question},
    {"role": "assistant", 
    "content": response_1},
    {"role": "user", 
    "content": question},
    {"role": "assistant", 
    "content": response_2},
    {"role": "user", 
    "content": question},
    {"role": "assistant", 
    "content": response_3},
    {"role": "user", 
    "content": question}
]
\end{lstlisting}
\end{small}

\subsection{TruthfulQA: Response With Reflection}
\begin{small}
\begin{lstlisting}
messages=[
    {"role": "system", 
    "content": "You are a helpful 
    assistant."},
    {"role": "user", 
    "content": question},
    {"role": "assistant", 
    "content": response_1},
    {"role": "user", 
    "content": "Please critique your
    responses"},
    {"role": "assistant", 
    "content": critique_1},
    {"role": "user", 
    "content": question},
    {"role": "assistant", 
    "content": response_2},
    {"role": "user", 
    "content": "Please critique your
    responses"},
    {"role": "assistant", 
    "content": critique_2},
    {"role": "user", 
    "content": question},
    {"role": "assistant", 
    "content": response_3},
    {"role": "user", 
    "content": "Please critique your
    responses"},
    {"role": "assistant", 
    "content": critique_3},
    {"role": "user", 
    "content": question}
]
\end{lstlisting}
\end{small}

\subsection{HotpotQA: Standard Prompt}
\begin{small}
\begin{lstlisting}
messages=[
    {"role": "system", 
    "content": "You are a helpful 
    assistant. Answer the question 
    based on the context provided. 
    Provide extremely concise answers 
    with no explanation."},
    {"role": "user", 
    "content": "Context: Earth: The
    Earth is the third planet from 
    the Sun. Question: Which planet 
    is Earth from the Sun? Answer: 
    Third"},
    {"role": "user", 
    "content": f"Context: 
    {formatted_context}\n
    Question: {question}\nProvide a 
    short answer without 
    explanation."}
]
\end{lstlisting}
\end{small}

\subsection{HotpotQA: Response Critique
Prompt}
\begin{small}
\begin{lstlisting}
messages=[
    {"role": "system", 
    "content": "You are a helpful 
    assistant. Answer the question 
    based on the context provided."},
    {"role": "user", 
    "content": f"Context: 
    {formatted_context}\n
    Question: {question}"},
    {"role": "assistant", 
    "content": f"{response}"},
    {"role": "user", 
    "content": f"Please review and 
    critique your previous response, 
    and keep in mind not to add any 
    unnecessary apologies. You can 
    refer back to the original 
    context if needed."}
]
\end{lstlisting}
\end{small}

\subsection{HotpotQA: Response Without
Reflection}
\begin{small}
\begin{lstlisting}
messages=[
    {"role": "system", 
    "content": "You are a helpful 
    assistant. Answer the question 
    based on the context provided. 
    Provide extremely concise answers 
    with no explanation."},
    {"role": "user", 
    "content": "Context: Earth: The 
    Earth is the third planet from 
    the Sun. Question: Which planet 
    is Earth from the Sun? 
    Answer: Third"},
    {"role": "user", 
    "content": f"Context: 
    {formatted_context}\n
    Question: {question}\n
    Provide a short answer without 
    explanation."},
    {"role": "assistant", 
    "content": f"{response_1}"},
    {"role": "user", 
    "content": f"{question}\n
    Provide a short answer without
    explanation."},
    {"role": "assistant", 
    "content": f"{response_2}"},
    {"role": "user", 
    "content": f"{question}\n
    Provide a short answer without 
    explanation."},
    {"role": "assistant", 
    "content": f"{response_3}"},
    {"role": "user", 
    "content": f"{question}\n
    Provide a short answer without
    explanation."},
    {"role": "assistant", 
    "content": f"{response_4}"},
    {"role": "user", 
    "content": f"{question}\n
    Provide a short answer without
    explanation."},
]
\end{lstlisting}
\end{small}

\subsection{HotpotQA: Response With
Reflection}
\begin{small}
\begin{lstlisting}
messages=[
    {"role": "system", 
    "content": "You are a helpful 
    assistant. Answer the question 
    based on the context provided. 
    Provide extremely concise answers 
    with no explanation."},
    {"role": "user", 
    "content": "Context: Earth: The 
    Earth is the third planet from the 
    Sun. Question: Which planet is Earth 
    from the Sun? Answer: Third"},
    {"role": "user", 
    "content": f"Context: 
    {formatted_context}\n
    Question: {question}\n
    Provide a short answer without 
    explanation."},
    {"role": "assistant", 
    "content": f"{response_1}"},
    {"role": "user", 
    "content": f"Please review and 
    critique your previous response, 
    and keep in mind not to add any 
    unnecessary apologies. You can 
    refer back to the original context 
    if needed."},
    {"role": "assistant", 
    "content": f"{critique_1}"},
    {"role": "user", 
    "content": f"{question}\n
    Provide a short answer without
    explanation."},
    {"role": "assistant", 
    "content": f"{response_2}"},
    {"role": "user", 
    "content": f"Please review and 
    critique your previous response, 
    and keep in mind not to add any 
    unnecessary apologies. You can 
    refer back to the original context 
    if needed."},
    {"role": "assistant", 
    "content": f"{critique_2}"},
    {"role": "user", 
    "content": f"{question}\n
    Provide a short answer without 
    explanation."},
    {"role": "assistant", 
    "content": f"{response_3}"},
    {"role": "user", 
    "content": f"Please review and 
    critique your previous response, 
    and keep in mind not to add any 
    unnecessary apologies. You can 
    refer back to the original 
    context if needed."},
    {"role": "assistant", 
    "content": f"{critique_3}"},
    {"role": "user", 
    "content": f"{question}\n
    Provide a short answer without 
    explanation."},
    {"role": "assistant", 
    "content": f"{response_4}"},
    {"role": "user", 
    "content": f"Please review and 
    critique your previous response, 
    and keep in mind not to add any 
    unnecessary apologies. You can 
    refer back to the original context 
    if needed."},
    {"role": "assistant", 
    "content": f"{critique_4}"},
    {"role": "user", 
    "content": f"{question}\n
    Provide a short answer without
    explanation."}
]
\end{lstlisting}
\end{small}
\subsection{HotpotQA: Fake Evidence Generation}
\begin{small}
\begin{lstlisting}
messages=[
    {"role": "system", 
    "content": "You are a helpful 
    assistant."},
    {"role": "user", 
    "content": f"Here is a question: 
    {question}. Please create 10 
    different versions of 'fake 
    supporting facts' based on the 
    following real supporting facts. 
    Modify only one sentence in each 
    version, making sure the modified 
    sentence is still relevant but 
    contains false information. Keep 
    the other sentences unmodified. 
    Each version of fake supporting 
    facts should have the same number 
    of sentences as the real 
    supporting facts."},
    {"role": "user", 
    "content": f"Real Supporting 
    Facts:{real_sf}"},
    {"role": "user", 
    "content": "Please generate the 
    fake supporting facts versions. 
    Remember to index all the sentences. 
    You must generate 10 versions 
    before you stop."},
    {"role": "user", 
    "content": 
    f"Fake Supporting Facts Version 1:\n
    [Insert manipulated sentences here]\n
    Fake Supporting Facts Version 2:\n
    [Insert manipulated sentences here]\n
    Fake Supporting Facts Version 3:\n
    [Insert manipulated sentences here]\n
    Fake Supporting Facts Version 4:\n
    [Insert manipulated sentences here]\n
    Fake Supporting Facts Version 5:\n
    [Insert manipulated sentences here]\n
    Fake Supporting Facts Version 6:\n
    [Insert manipulated sentences here]\n
    Fake Supporting Facts Version 7:\n
    [Insert manipulated sentences here]\n
    Fake Supporting Facts Version 8:\n
    [Insert manipulated sentences here]\n
    Fake Supporting Facts Version 9:\n
    [Insert manipulated sentences here]\n
    Fake Supporting Facts Version 10:\n
    [Insert manipulated sentences here]"},
]
\end{lstlisting}
\end{small}

\section{Illustration of API Calling Processes}
\label{app:api_call_example}
In this section, we provide a simple example to illustrate the API calling process under our \protocol~, conditional prompting and the Exploration-Only Baseline. 

\subsection{\protocol~ API Calling Process}
\begin{small}
\begin{lstlisting}
(Splitters and other special tokens are omitted)
*First API Call*:
[Instructions and Context]
Question: [question]
Response: ____ (Sample response_1, response_2 here.)

*Second API Call*:
[Instructions and Context]
Question: [question]
Response: response_1
[Instruction for Reflection]
Reflection: ____ (Sample reflection_1 here.)

*Third API Call*:
[Instructions and Context]
Question: [question]
Response: response_2 
[Instruction for Reflection]
Reflection: ____ (Sample reflection_2 here.)

*Final API Call (to get the final revised answer)*:
[Instructions and Context]
Question: [question]
Response: response_1
[Instruction for Reflection]
Reflection: reflection_1

Question: [question]
Response: response_2
[Instruction for Reflection]
Reflection:  reflection_2

Question: [question]
Response:  ____ (Sample final_response here)
\end{lstlisting}
\end{small}

\subsection{Conditional Prompting Baseline API Calling Process}
\begin{small}
\begin{lstlisting}
*First API Call*:
[Instructions and Context]
Question: [question]
Response: ____ (sample response_1 here)

*Second API Call*:
[Instructions and Context]
Question: [question]
Response: response_1
[Instruction for Reflection]
Reflection: ____ (sample reflection_1 here)

*Third API Call*:
[Instructions and Context]
Question: [question]
Response: response_1
[Instruction for Reflection]
Reflection: reflection_1

Question: [question]
Response: ___ (sample response_2 here)

...

*Final API Call (to get the final revised answer)*:
[Instructions and Context]
Question: [question]
Response: response_1
[Instruction for Reflection]
Reflection: reflection_1

Question: [question]
Response: response_2
[Instruction for Reflection]
Reflection:  reflection_2

Question: [question]
Response:  final_reponse
\end{lstlisting}
\end{small}

\subsection{Exploration-Only Baseline API Calling Process}
\begin{small}
\begin{lstlisting}
*First API Call*:

[Instructions and Context]
Question: [question]
Response: ____ (sample response_1, response_2 here)

*Final API Call*:

[Instructions and Context]
Question: [question]
Response: response_1
Question: [question]
Response: response_2 
Question: [question]
Response: ____ (sample final_response here)
\end{lstlisting}
\end{small}

\section{Challenges in Predicting the Correctness Margin for Model Comprehension}
\label{app:challenge_predict_correctness_margin}

The effectiveness of a model's self-reflection largely hinges on its "correctness margin," a metric quantifying its understanding of both the question and its context. Ideally, we would like to predict this margin through user prompts, thereby allowing the user to make an informed decision on whether to enable the model's self-reflection capability.

Nevertheless, our experiments indicate that current models struggle to self-assess their understanding reliably. Below, we outline our prompt design used for this experiment:

\begin{small}
\begin{lstlisting}
messages=[
    {"role": "system", 
    "content": "You are a helpful
    assistant. Answer the question based 
    on the context provided. Provide 
    extremely concise answers with no 
    explanation."},
    {"role": "user", 
    "content": f"Context:
    {formatted_context}\n
    Question: {question}"},
    {"role": "assistant", 
    "content": f"{response}"},
    {"role": "user", 
    "content": "\nYou have just answered 
    a question. Now, please evaluate your 
    own comprehension of the question and 
    answer provided. Rate your level of 
    understanding on a scale from -5 to 5. 
    A rating of 5 signifies extreme 
    certainty that you understand the 
    question, while a rating of -5 
    indicates extreme uncertainty or lack 
    of understanding."},
]
\end{lstlisting}
\end{small}

We tested this prompt structure on two sets of questions: one where all $10$ model responses were incorrect, and another where all $10$ were correct. If the model were capable of accurately evaluating its own comprehension, it should consistently rate its understanding at $-5$ for questions in the all-wrong dataset and $5$ for those in the all-right dataset. However, after experimenting with $20$ examples from each dataset, we found that the model consistently assigned high scores (typically $4$ or $5$) regardless of the dataset origin. Thus, reliable self-assessment remains an open challenge for current models.

\section{Scientific Artifacts}
\label{app:scientific_artifacts}

In this paper, we use the following artifacts:
\begin{itemize}
    \item TruthfulQA \cite{lin2022truthfulqa} is a benchmark assessing a language model's ability to generate truthful answers for 817 diverse questions in 38 categories, requiring models to avoid false answers commonly found in human texts due to misconceptions or false beliefs. We use it for the preliminary studies on reflective thinking in LLMs. It is licensed under the Apache License, Version 2.0.
    \item HotpotQA \cite{yang2018hotpotqa} is a 113k question-answer dataset based on Wikipedia that requires multi-document reasoning, features diverse questions unconstrained by knowledge bases or schemas, provides sentence-level supporting facts for strong supervision and explanation, and introduces a new factoid comparison question type to evaluate QA systems' extraction and comparison abilities. We use it for evaluating reflective thinking in LLMs. It is distributed under a CC BY-SA 4.0 License.
    \item openai-python\footnote{https://github.com/openai/openai-python} (v0.27.8) provides convenient access to the OpenAI REST API from any Python 3.7+ application. We use it to access ChatGPT models. It is licensed under the Apache License, Version 2.0.

\end{itemize}

\section{Results on LLaMA-2-7b-chat}
\label{app:llama2_results}
We extend our experiments to the open-sourced model LLaMA-2-7b-chat~\cite{touvron2023llama}, and the results support the conclusions that we draw from our experiments on ChatGPT, indicating that our findings can be generalized to different models.

\begin{figure}{}
\centering
\small
\includegraphics[width=0.4\textwidth]{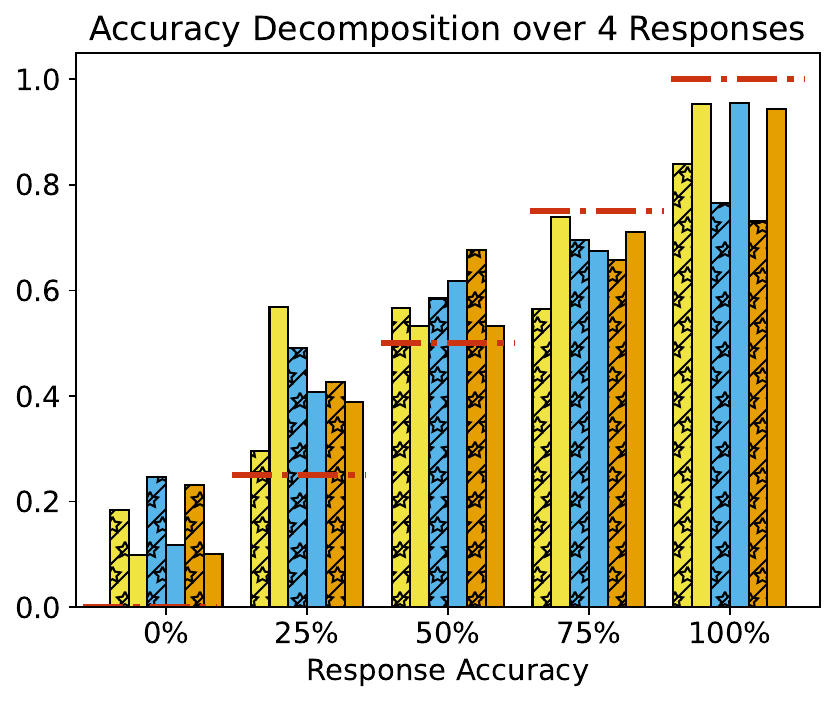}
  \includegraphics[width=0.4\textwidth]{figures/legend.pdf}
  \vspace{-2mm}
  \caption{Performance Decomposition on Question Difficulty and Response Accuracy (LLaMA-2-chat). 
  }

  \label{fig:perf_decomp_4response_natural_llama}
\end{figure}

More specifically, for the preliminary study on performance across TruthfulQA and HotpotQA, the results on LLaMA-2-7b-chat (see \cref{tab: merged_res_llama2}) are consistent with the results obtained from ChatGPT (see \cref{tab: merged_res}): self-reflection prompts improve performance on TruthfulQA while worsening performance on HotpotQA. Additionally, we conduct our error analysis on the results of HotpotQA, breaking down model performance based on levels of question difficulty and model comprehension. The LLaMA-2-7b-chat results for this analysis (\cref{fig:perf_decomp_4response_natural_llama}) also closely follow the trend observed in the results from ChatGPT (\cref{fig:perf_decomp_4response_natural}): self-reflection is more beneficial when the model's initial responses are incorrect and when the question difficulty is higher.

We do not replicate the 10-response artificial experiments on LLaMA-2-7b-chat due to the context length limit. The context length for LLaMA-2 is 4096, which is shorter than our context length for the task. We replicate this experiment in \cref{app:mixtral_results} as Mixtral models have larger token limits.

\begin{table}[t!]
\centering
\resizebox{0.49\textwidth}{!}{%
\begin{tabular}{@{}p{2cm}p{2cm}p{2cm}p{2cm}@{}}
\toprule
Metric & {Standard Prompting} & Exploration-Only & Self-Reflection \\
\midrule
\multicolumn{4}{c}{TruthfulQA} \\
Rouge-1 & $53.8\pm 0.4 $ & $51.7$ & $\textbf{53.8}$ \\
BLEURT & $ 60.9 \pm 0.6$ & $58.2$ & $\textbf{63.0}$ \\
\midrule
\multicolumn{4}{c}{HotpotQA} \\
Accuracy* & $ 61.0\pm 1.0$ & $\textbf{62.9}$ & $57.5$ \\
\bottomrule
\end{tabular}
}
\caption{Self-reflection \protocol{} experiment results on QA datasets using LLaMA-2-chat. Bold-facing indicates the best-performing method under each metric. *Evaluated manually. 
}
\label{tab: merged_res_llama2}
\vspace{-5pt}
\end{table}

\begin{table}[t!]
\centering
\resizebox{0.49\textwidth}{!}{%
\begin{tabular}{@{}p{2cm}p{2cm}p{2cm}p{2cm}@{}}
\toprule
Metric & {Standard Prompting} & Exploration-Only & Self-Reflection \\
\midrule
\multicolumn{4}{c}{TruthfulQA} \\
Rouge-1 & $59.1\pm 1.0 $ & $61.3$ & $\textbf{63.3}$ \\
BLEURT & $ 71.5 \pm 0.4$ & $\textbf{73.9}$ & $71.7$ \\
\midrule
\multicolumn{4}{c}{HotpotQA} \\
Accuracy* & $ 89.8 \pm 0.3$ & $ \textbf{90.9}$ & $89.2$ \\
\bottomrule
\end{tabular}
}
\caption{Self-reflection \protocol{} experiment results on QA datasets using Mixtral-8x7B-v0.1. Bold-facing indicates the best-performing method under each metric. *Evaluated manually. 
}
\label{tab: merged_res_mixtral}
\vspace{-5pt}
\end{table}

\section{Results on Mixtral-8x7B-v0.1}
\label{app:mixtral_results}

\begin{figure}{}
\centering
\small
\includegraphics[width=0.4\textwidth]{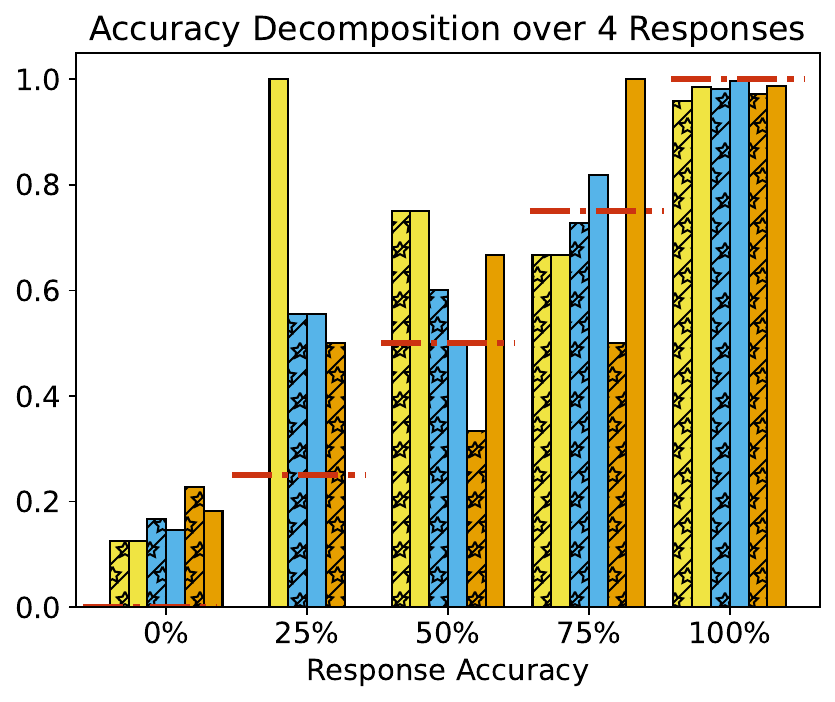}
  \includegraphics[width=0.4\textwidth]{figures/legend.pdf}
  \vspace{-2mm}
  \caption{Performance Decomposition on Question Difficulty and Response Accuracy (Mixtral-8x7B-v0.1). 
  }

  \label{fig:perf_decomp_4response_natural_mixtral}
\end{figure}

\begin{figure}{}
\centering
\small
\includegraphics[width=0.48\textwidth]{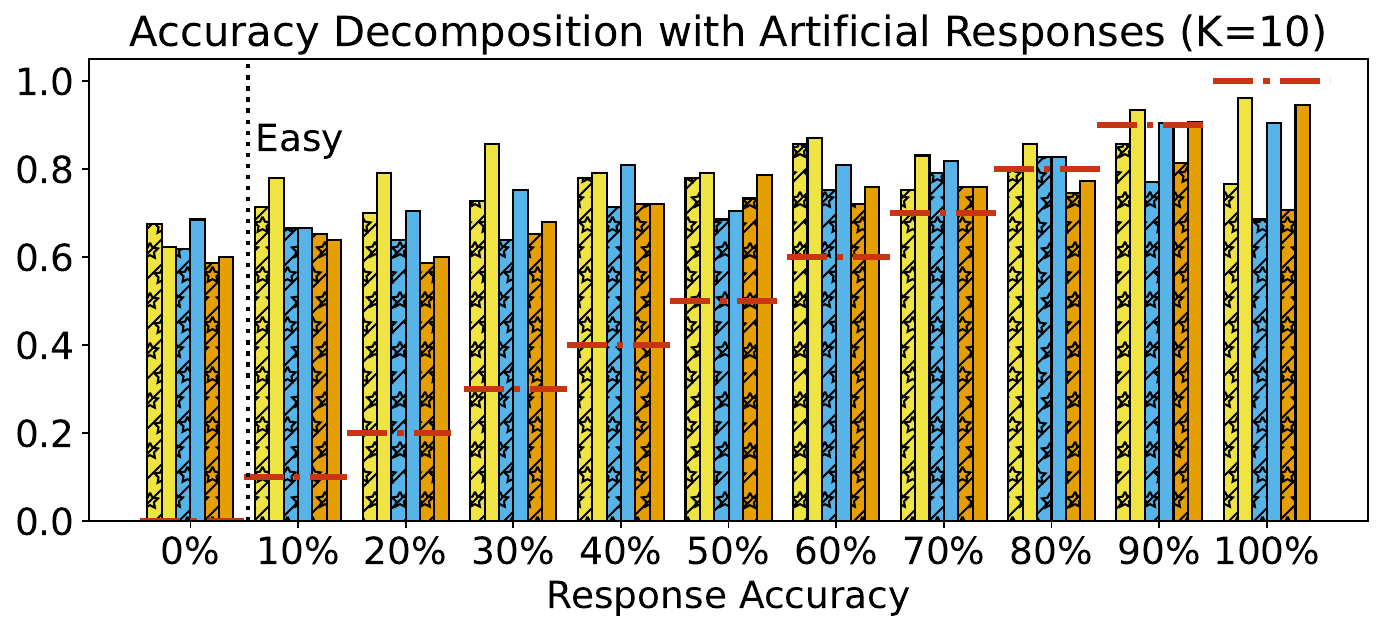}
  \includegraphics[width=0.4\textwidth]{figures/legend.pdf}
  \vspace{-2mm}
  \caption{Performance Decomposition on Question Difficulty and Response Accuracy (Artificial Responses) for Mixtral-8x7B-v0.1.  
  }

  \label{fig:perf_decomp_10response_natural_mixtral}
\end{figure}

We repeat our experiments on Mixtral-8x7B-v0.1. For the preliminary study on performance across TruthfulQA and HotpotQA (\cref{tab: merged_res_mixtral}, we again observe a similar trend to ChatGPT: while self-reflection may help improve the performance on TruthfulQA, it harms the performance on HotpotQA. Then, we again break down model performance on HotpotQA based on question difficulty levels and model comprehension in \cref{fig:perf_decomp_4response_natural_mixtral}. Here we observe a somewhat different pattern: under all question difficulty levels and model comprehension, self-reflection prompting fails to improve the performance. To further verify this finding, we also conduct the artificial response experiments, with results in \cref{fig:perf_decomp_10response_natural_mixtral}. Here we see that self-reflection prompting is not always harmful to performance (e.g., under $0\%$ RA, self-reflection helps improve the performance of easy questions.), but in most cases it is harmful.

\begin{figure}{}
\centering
\small
\includegraphics[width=0.48\textwidth]{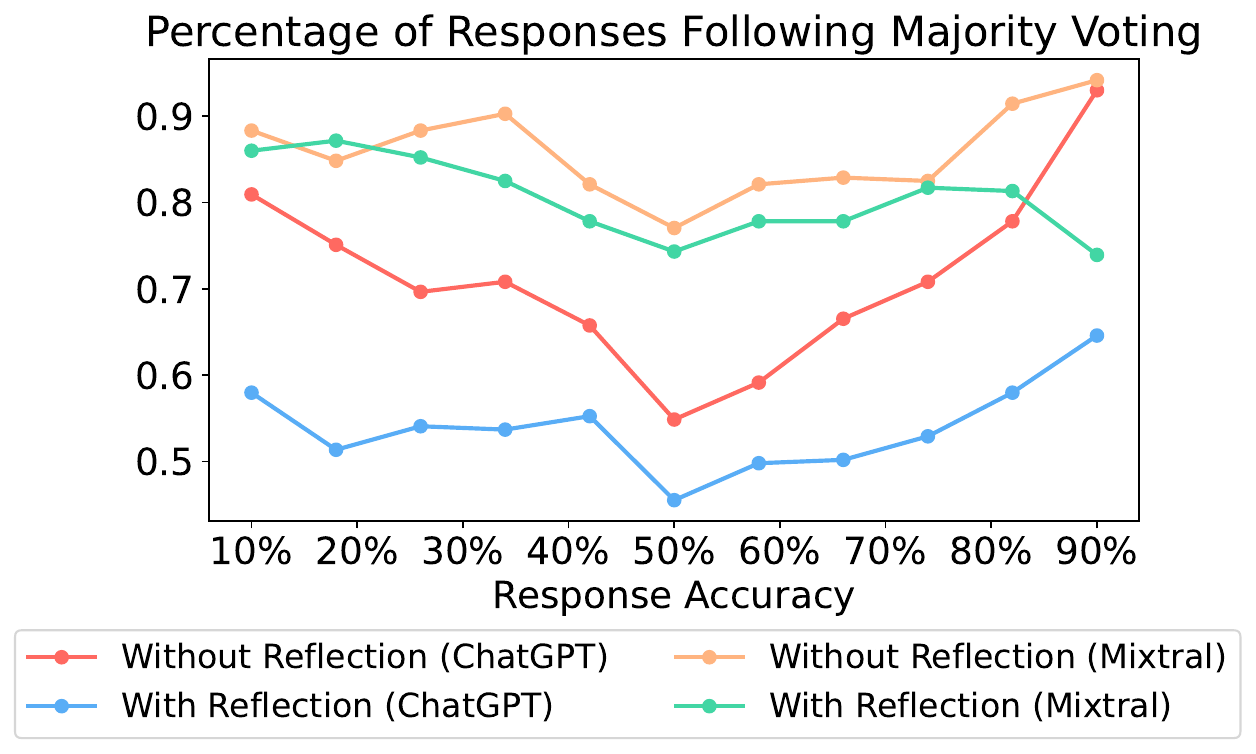}
  \caption{Majority Voting Analysis (Mixtral-8x7B-v0.1). 
  }

  \label{fig:majority_voting_mixtral}
\end{figure}

We hypothesize that these divergent patterns arise because this particular model may be less sensitive in general to instructions for reflection, or less well-equipped to understand them, such that the reflection part serves mostly as a distractor in the input. We examine this hypothesis in the majority voting experiments (\cref{fig:majority_voting_mixtral}) and find that compared with ChatGPT, the addition of self-reflection exerts minimal impact on majority voting trends, suggesting that it is comparatively difficult to use self-reflection prompting to change the default behaviors in the case of this model. This is consistent with our hypothesis that Mixtral-8x7B-v0.1 lacks sensitivity or competence in self-reflection, so we speculate that additional training may be needed to help this model to unlock self-reflection prompting potential.

\end{document}